\documentclass[runningheads]{llncs}
\usepackage{graphicx}
\usepackage{xspace}
\usepackage{epsfig}
\usepackage{amsmath}
\usepackage{amssymb}
\usepackage{acro}
\usepackage{multirow}
\usepackage{booktabs}
\usepackage{makecell}
\usepackage{xcolor}
\usepackage[colorlinks, linkcolor=blue, anchorcolor=blue, citecolor=blue]{hyperref}
\newcommand{\ie}{\emph{i.e.}, }
\newcommand{\eg}{\emph{e.g.}, }

\newcommand{\etal}{\emph{et al.}}

\DeclareAcronym{PACS}{
short=PACS,
long=picture archiving and communication system
}

\DeclareAcronym{SSL}{
short=SSL,
long=semi-supervised learning
}

\DeclareAcronym{CNN}{
short=CNN,
long=convolutional neural network
}

\DeclareAcronym{GCN}{
short=GCN,
long=graph convolutional network
}

\DeclareAcronym{DAG}{
short=DAG,
long=deep adaptive graph
}

\DeclareAcronym{GT}{
short=GT,
long=ground truth
}

\DeclareAcronym{JS}{
short=JS,
long=Jensen–Shannon
}

\DeclareAcronym{EMA}{
short=EMA,
long=exponential moving average
}

\begin{document}
\title{Scalable Semi-supervised Landmark Localization for X-ray Images using Few-shot Deep Adaptive Graph}
\titlerunning{Landmark Localization in X-rays using Few-shot Deep Adaptive Graph}
%
\author{Xiao-Yun Zhou \inst{1}, Bolin Lai \inst{2}, Weijian Li \inst{3}, Yirui Wang \inst{1}, Kang Zheng \inst{1}, Fakai Wang \inst{4}, Chihung Lin \inst{5}, Le Lu \inst{1}, Lingyun Huang \inst{2}, Mei Han \inst{1}, Guotong Xie \inst{2}, Jing Xiao \inst{2}, Kuo Chang-Fu \inst{5}, Adam Harrison \inst{1}, Shun Miao \inst{1}}
\authorrunning{Paper ID 591}
%

\institute{PAII Inc., Bethesda, MD, USA \and Ping An Technology, Shenzhen, China \and University of Rochester, Rochester, NY, USA \and University of Maryland, College Park, MD, USA \and Chang Gung Memorial Hospital, Linkou, Taiwan, ROC}
\maketitle              
\begin{abstract}
Landmark localization plays an important role in medical image analysis. Learning based methods, including \ac{CNN} and \ac{GCN}, have demonstrated the state-of-the-art performance. However, most of these methods are fully-supervised and heavily rely on manual labeling of a large training dataset. In this paper, based on a fully-supervised graph-based method, \ac{DAG}, we proposed a semi-supervised extension of it, termed few-shot \ac{DAG}, \ie five-shot \ac{DAG}. It first trains a \ac{DAG} model on the labeled data and then fine-tunes the pre-trained model on the unlabeled data with a teacher-student \ac{SSL} mechanism. In addition to the semi-supervised loss, we propose another loss using \ac{JS} divergence to regulate the consistency of the intermediate feature maps. We extensively evaluated our method on pelvis, hand and chest landmark detection tasks. Our experiment results demonstrate consistent and significant improvements over previous methods.

\end{abstract}
\acresetall
\section{Introduction}
\label{sec:intro}

Landmark localization is a fundamental tool for a wide spectrum of medical image analysis applications, including image registration~\cite{han2015robust}, developmental dysplasia diagnosis~\cite{liu2020misshapen}, and scoliosis assessment~\cite{yi2020vertebra}. 
Although many recent improvements have been proposed~\cite{juneja2021review,wu2019facial}, most of them are still based on fully-supervised learning and rely heavily on the manual labeling of a large amount of training data. Human labeling is prohibitively expensive and requires medical expertise. Thus, it is challenging to obtain large-scale labeled training data in practical applications and decreased scales of training data can impede achieving strong performance. Yet, large-scale unlabeled X-ray images can be efficiently collected from \acp{PACS}. Hence, a promising strategy is to adopt \ac{SSL} scheme, which enables efficient learning from both labeled and unlabeled data. 

State-of-the-art landmark detection methods are typically learning-based, \ie \ac{GCN}~\cite{lu2020contour,lu2020learning,li2020structured}, heatmap regression~\cite{wu2018look,zhu2019robust,sun2019deep,valle2018deeply}, and coordinate regression~\cite{lv2017deep,yu2016deep,zhang2015learning,trigeorgis2016mnemonic}. Among these, \ac{DAG}~\cite{li2020structured} employs \acp{GCN} to exploit both the visual and structural information to localize landmarks. By incorporating a shape prior, \ac{DAG} reaches a higher robustness compared to heatmap-based and coordinate regression-based methods~\cite{li2020structured}.
While there are efforts towards \ac{SSL} for landmark localization~\cite{honari2018improving,tang2018facial}, they are based off of \ac{CNN}-only approaches, rather than the state-of-the-art \ac{GCN}-based \ac{DAG}. Some other prominent \ac{SSL} successes in medical imaging have also been reported, such as for  segmentation~\cite{cui2019semi,raju2020co} and abnormality detection~\cite{wang2020knowledge}, but these are also \ac{CNN}-based. Most other \ac{SSL} methods are mainly developed for natural image classification tasks~\cite{laine2016temporal,lee2013pseudo,berthelot2019mixmatch,tarvainen2017mean}, which likewise do not address \ac{GCN}-based landmark localization. 

\begin{figure}[t]
    \centering
    \includegraphics[width=1.0\textwidth]{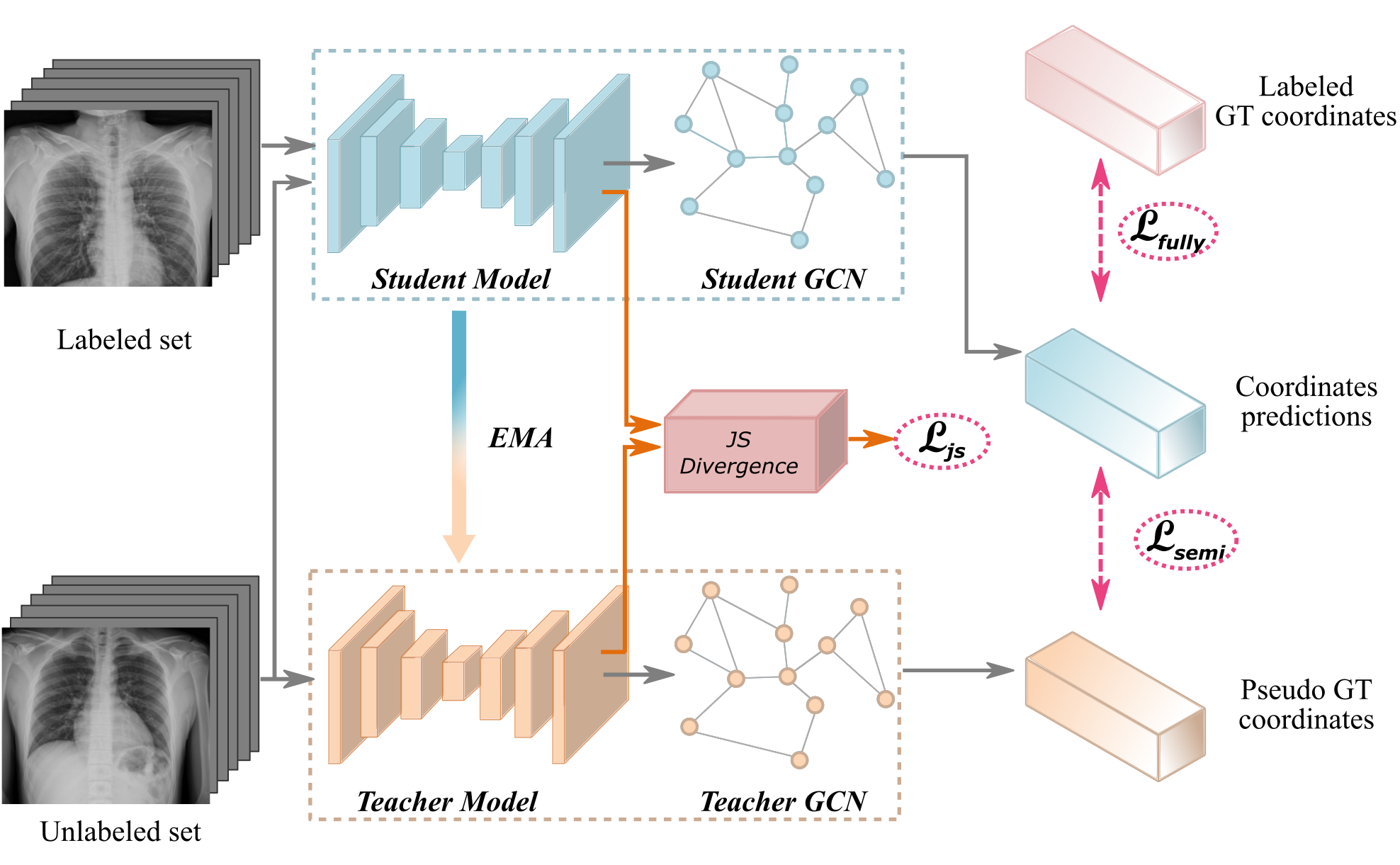}
    \caption{An illustration of the proposed few-shot DAG framework with the teacher student \ac{SSL} scheme and \ac{JS} divergence loss.}
    \label{fig:fewshotdag}
\end{figure}

In this work, we propose  few-shot \ac{DAG}, an effective \ac{SSL} approch for landmark detection. Few-shot \ac{DAG} can achieve strong landmark localization performance with only a few training images (\eg five). The framework of few-shot \ac{DAG} is illustrated in Fig.~\ref{fig:fewshotdag}. We first train a fully-supervised \ac{DAG} model on the labeled data and then fine-tune the pre-trained \ac{DAG} model using \ac{SSL} on the unlabeled data. Inspired by~\cite{tarvainen2017mean}, for \ac{SSL}, dual models are used, \ie{} a teacher and a student model. The output of the teacher model is used as the pseudo \ac{GT} to supervise the training and back-propagation of the student model. The parameters of the teacher model are updated by the \ac{EMA} of the parameters of the student model. In addition to the semi-supervised loss inspired by~\cite{tarvainen2017mean}, we further add a \ac{JS} divergence loss on the intermediate feature map, to encourage similar feature distributions between the teacher and student models. The proposed few-shot \ac{DAG} is validated on pelvis, hand and chest X-ray images with 10, 10, and 20 labeled samples and 5000, 3000, and 5000 unlabeled samples, showing consistent, notable and stable improvements compared with state-of-the-art fully-supervised methods~\cite{payer2019integrating} and other semi-supervised methods~\cite{sohn2020simple,laine2016temporal}.

\section{Method}
\label{sec:method}

Our work enhances prior efforts at using \ac{DAG} for landmark detection~\cite{li2020structured}. We first briefly describe the network structure and training mechanism of DAG in Sec.~\ref{sec:dag} and then introduce the proposed SSL extension of DAG with the \ac{JS} divergence loss in Sec.~\ref{sec:ssl}.

\subsection{Deep Adaptive Graph}
\label{sec:dag}

\ac{DAG}~\cite{li2020structured} formulates landmark localization as a graph evolution task, where the vertices of the graph represent the landmarks to be localized. The evolution starts from the mean shape generated from the training data, and the evolution policy is modeled by a \ac{CNN} feature extractor followed by two \ac{GCN}s. The \ac{CNN} encodes the input image as a feature map, from which graph features are extracted via bi-linear interpolation at the vertex locations. The graph with features is further processed by cascading a global \ac{GCN} and local \acp{GCN} to respectively estimate the affine transformation and vertex displacements toward the targets.

The DAG is trained via fully-supervised learning using a global \ac{GCN} and local \ac{GCN} loss. Specifically, the global \ac{GCN} outputs an affine transformation that globally transforms the initial graph vertices (\ie the mean shape). The average L1 distances between the affine transformed vertices and the \ac{GT} locations are calculated as the global loss:
\begin{equation}
	\mathcal{L}_{global}= \left[\mathbb{E} \left( |\mathbf{v}_{global}-\mathbf{v}_{gt}|\right)-m\right]_+, \label{eqn:global_loss}
\end{equation}
where $[x]_+ := max(0, x)$, $\mathbf{v}_{global}$ and $\mathbf{v}_{gt}$ denote the affine transformed and \ac{GT} vertices, respectively. $m$ is a hyper-parameter specifying the margin of allowable error.
The local \ac{GCN} iteratively displaces $\mathbf{v}_{global}$ to refine their locations. The average L1 distances between the displaced vertices and the \ac{GT} are calculated as the local loss:
\begin{equation}
	\mathcal{L}_{local}= \mathbb{E} \left( |\mathbf{v}_{local}-\mathbf{v}_{gt}| \right), \label{eqn:local_loss}
\end{equation}
where $\mathbf{v}_{local}$ denotes the vertices after displacement by the local \ac{GCN}. The final loss for \ac{DAG} is $\mathcal{L}_{global} + w_1 \times \mathcal{L}_{local}$, where $w_1$ is a weight used to adjust the ratio between the global and local loss.

\subsection{Few-shot \ac{DAG}}
\label{sec:ssl}

Inspired by \cite{tarvainen2017mean}, we adopt a mean teacher mechanism to exploit both the labeled and unlabeled data. In particular, we first train a \ac{DAG} model using only the labeled dataset, referred to as the \textit{pre-trained model}. In the mean teacher training, the teacher and student models share the same architecture and are both initialized using the \textit{pre-trained model}. The same input images are fed into the teacher and student models. Gaussian noises are added to the input images of the student model as an additional augmentation. For unlabeled images, a consistency loss is enforced between the teacher and student models. In the proposed few-shot \ac{DAG}, we apply two forms of unlabeled loss.

First, the landmarks detected by the teacher model are used as the pseudo \ac{GT} to supervise the training of the student model. In particular, the output $\mathbf{v}_{local}$ of the teacher model is used as the pseudo GT for the student model to calculate the global and local losses of Equ. \eqref{eqn:global_loss} and \eqref{eqn:local_loss}, respectively. While this is helpful, it only applies a sparse consistency constraint on the \ac{GCN} outputs. As a result, we apply a second loss in the form of \ac{JS} divergence between the \ac{CNN} feature maps of the teacher and student model, encouraging a similar distribution between the two. Specifically, the output feature maps of \ac{CNN} are converted to pseudo-probabilities via a Softmax along the channel dimension. The \ac{JS} divergence loss is then formulated as:
\begin{align}
    \mathcal{L}_{js} = \frac{1}{2|\Omega|}  \sum_{x \in \Omega}\left(D(\mathbf{a}_{S}(x), \mathbf{m}(x)) + D(\mathbf{a}_{T}(x), \mathbf{m}(x)\right) 
\label{equ:offset}
\end{align}
where $D(.)$ is the  Kullback–Leibler divergence, $\mathbf{a}_{S}$ and $\mathbf{a}_{T}$ are the student and teacher activation maps, respectively, $\Omega$ is their batch, spatial and channel domain, and $\mathbf{m}$ is the mean of $\mathbf{a}_{S}$ and $\mathbf{a}_{T}$.

For the labeled data, we use the fully-supervised loss: $\mathcal{L}_{global} + w_1 \times \mathcal{L}_{local}$. For the unlabeled data, we calculate $\mathcal{L'}_{global} + w_1 \times \mathcal{L'}_{local} + w_2*\mathcal{L}_{js}$, where $\mathcal{L'}_{global} + w_1 \times \mathcal{L'}_{local}$ use the pseudo GT produced by the teacher model and $w_2$
balances the contribution of the \ac{JS} divergence loss. The labeled and unlabeled batches are fed with a ratio $1:R$ (R is 100 in our experiments) to form the semi-supervised training iterations. Finally, the weights of the student model are updated through back-propagation of the loss. The weights of the teacher model are updated iteratively via the \ac{EMA} of the student model's weights~\cite{tarvainen2017mean}:
\begin{equation}
    \theta_{T}^t = \alpha \theta_{T}^{t-1} + (1-\alpha)\theta_S^{t},
\end{equation}
where $\theta_{T}$ and $\theta_S$ are the weights of the teacher and student models, respectively, $t$ is the training step, and $\alpha$ is a smoothing coefficient to control the pace of knowledge updates.

\section{Results}
\label{sec:experiments}

\begin{table}[h]	
	\setlength{\tabcolsep}{0.07cm} 
	\caption{The mean and std Euclidean error and the failure rate of the proposed method, with comparisons to Payer \etal, pseudo label, $\rm{\Pi}$-model and temporal ensemble on the pelvis, hand and chest datasets. Best performance is in bold.}
	\label{tab:comparison}
	\centering
	\begin{tabular}{c|c|c|c|c}
		\hline 
		Data & Method & Mean error & Std error & Failure rate \\
		\hline
		\multirow{7}{*}{Pelvis}& Payer \textit{et al.}~\cite{payer2019integrating} & 46.29 & 106.63 & 12.62\% \\
		\cline{2-5}
		& Pseudo label~\cite{sohn2020simple}& 20.50 & 34.27 & 2.21\% \\
		\cline{2-5}
		& $\rm{\Pi}$-Model~\cite{laine2016temporal}& 58.31 & 98.41 & 14.66\% \\
		\cline{2-5}
		& Temporal ensemble~\cite{laine2016temporal}& 21.12 & 42.90 & 2.07\% \\
		\cline{2-5}
		& DAG~\cite{li2020structured} & 25.89 & 44.60 & 4.29\% \\
		\cline{2-5}
		& Few-shot DAG & 19.63 & 34.29 & \textbf{1.27\%} \\
		\cline{2-5}
		& \textbf{\textit{Few-shot DAG + JS}} & \textbf{\textit{18.45}} & \textbf{\textit{30.69}} & \textit{1.31\%} \\
		
		\hline
		\multirow{7}{*}{Hand}& Payer \textit{et al.}~\cite{payer2019integrating} & 12.29 & 37.81 & 1.24\% \\
		\cline{2-5}
		& Pseudo label~\cite{sohn2020simple}& 9.27 & 24.82 & 0.77\% \\
		\cline{2-5}
		& $\rm{\Pi}$-Model~\cite{laine2016temporal}& 17.96 & 45.07 & 3.56\% \\
		\cline{2-5}
		& Temporal ensemble~\cite{laine2016temporal}& 10.20 & 22.35 & 0.81\% \\
		\cline{2-5}
		& DAG~\cite{li2020structured} & 10.97 & 27.60 & 1.51\% \\
		\cline{2-5}
		& Few-shot DAG & 9.07 & 21.76 & 0.50\% \\
		\cline{2-5}
		& \textbf{\textit{Few-shot DAG + JS}} & \textbf{\textit{9.07}} & \textbf{\textit{19.67}} & \textbf{\textit{0.47\%}} \\
		
		\hline
		\multirow{7}{*}{Chest}& Payer \textit{et al.}~\cite{payer2019integrating} & 61.41 & 131.27 & 5.75\% \\
		\cline{2-5}
		& Pseudo label~\cite{sohn2020simple}& 55.33 & 57.84 & 8.32\% \\
		\cline{2-5}
		& $\rm{\Pi}$-Model~\cite{laine2016temporal}& 208.38 & 138.45 & 64.80\% \\
		\cline{2-5}
		& Temporal ensemble~\cite{laine2016temporal}& 52.41 & 47.54 & 5.92\% \\
		\cline{2-5}
		& DAG~\cite{li2020structured} & 58.99 & 73.55 & 12.35\% \\
		\cline{2-5}
		& Few-shot DAG & 54.94 & 55.00 & 9.37\% \\
		\cline{2-5}
		& \textbf{\textit{Few-shot DAG + JS}} & \textbf{\textit{43.46}} & \textbf{\textit{47.22}} & \textbf{\textit{5.28\%}} \\

		\hline
	\end{tabular}
\end{table}

\subsubsection{Experimental setup}
The proposed few-shot DAG is validated on three X-ray data sets: pelvis (60 labeled images, 5000 unlabeled images, 6029 test images), hand (36 labeled images, 3000 unlabeled images, 93 test images), and chest (60 labeled images, 5000 unlabeled images, 1092 test images). All datasets were collected from \textit{anonymous hospital} after de-identification of the patient information. All experiments, except the ablation study, were conducted with 10/10, 10/6, and 20/10 training/validation examples for the pelvis, hand, and chest data respectively. For the ablation study, additional experiments with 1/5/50, 1/5/30, 1/5/10/50 training examples for the pelvis, hand and chest data are conducted to validate the scalability of the proposed few-shot \ac{DAG} on different numbers of training examples. The Euclidean distance between the \ac{GT} and the predicted landmarks is used as the main evaluation metric. In addition, the failure rate (defined as error larger than $5\%$ of the image width) is supplied as a supplementary evaluation metric. As the failure rate may change along the chosen threshold, we view the Euclidean error as more important and objective.

\begin{table}[h]	
	\setlength{\tabcolsep}{0.07cm} 
	\caption{The mean and std Euclidean error and the failure rate of \ac{DAG}, few-shot \ac{DAG} and few-shot \ac{DAG} + \ac{JS} on different numbers of training examples. Best performance is in bold. - indicates no convergence.}
	\label{tab:ablation}
	\centering
	\begin{tabular}{c|c|c|c|c|c}
		\hline 
		Data & Training samples & Method & Mean error & Std error & Failure rate \\
		\hline
		\multirow{10}{*}{Pelvis} & 1 & DAG~\cite{li2020structured} & - & - & - \\
		\cline{2-6}
		& \multirow{3}{*}{5} & DAG~\cite{li2020structured} & 55.53 & 80.05 & 14.72\% \\
		\cline{3-6}
		& & Few-shot DAG & 34.48 & 44.17 & 3.22\% \\
		\cline{3-6}
		& & \textbf{\textit{Few-shot DAG + JS}} & \textbf{\textit{27.31}} & \textbf{\textit{39.14}} & \textbf{\textit{3.19\%}} \\
		\cline{2-6}
		& \multirow{3}{*}{10} & DAG~\cite{li2020structured} & 25.89 & 44.60 & 4.29\% \\
		\cline{3-6}
		& & Few-shot DAG & 19.63 & 34.29 & \textbf{1.27\%} \\
		\cline{3-6}
		& & \textbf{\textit{Few-shot DAG + JS}} & \textbf{\textit{18.45}} & \textbf{\textit{30.69}} & \textit{1.31\%} \\
		\cline{2-6}
		& \multirow{3}{*}{50} & DAG~\cite{li2020structured} & 15.62 & 34.40 & 1.29\% \\
		\cline{3-6}
		& & Few-shot DAG & 13.44 & 30.03 & \textbf{0.55\%} \\
		\cline{3-6}
		& & \textbf{\textit{Few-shot DAG + JS}} & \textbf{\textit{13.37}} & \textbf{\textit{28.26}} & \textit{0.58\%} \\
		
		\hline
		\multirow{10}{*}{Hand} & 1 & DAG~\cite{li2020structured} & - & - & - \\
		\cline{2-6}
		& \multirow{3}{*}{5} & DAG~\cite{li2020structured} & 24.17 & 47.05 & 5.07\% \\
		\cline{3-6}
		& & Few-shot DAG & 23.30 & 36.13 & 2.52\% \\
		\cline{3-6}
		& & \textbf{\textit{Few-shot DAG + JS}} & \textbf{\textit{15.41}} & \textbf{\textit{31.99}} & \textbf{\textit{1.78\%}} \\
		\cline{2-6}
		& \multirow{3}{*}{10} & DAG~\cite{li2020structured} & 10.97 & 27.60 & 1.51\% \\
		\cline{3-6}
		& & Few-shot DAG & 9.07 & 21.76 & 0.50\% \\
		\cline{3-6}
		& & \textbf{\textit{Few-shot DAG + JS}} & \textbf{\textit{9.07}} & \textbf{\textit{19.67}} & \textbf{\textit{0.47\%}} \\
		\cline{2-6}
		& \multirow{3}{*}{50} & DAG~\cite{li2020structured} & 8.44 & 22.00 & 0.67\% \\
		\cline{3-6}
		& & Few-shot DAG & 8.09 & \textbf{17.51} & \textbf{0.40\%} \\
		\cline{3-6}
		& & \textbf{\textit{Few-shot DAG + JS}} & \textbf{\textit{7.74}} & \textit{17.55} & \textit{0.43\%} \\
		
		\hline
		\multirow{10}{*}{Chest} & 1 \& 5 & DAG~\cite{li2020structured} & - & - & - \\
		\cline{2-6}
		& \multirow{3}{*}{10} & DAG~\cite{li2020structured} & 133.09 & 121.57 & 38.49\% \\
		\cline{3-6}
		& & Few-shot DAG & 128.74 & 102.61 & 39.73\% \\
		\cline{3-6}
		& & \textbf{\textit{Few-shot DAG + JS}} & \textbf{\textit{76.11}} & \textbf{\textit{78.01}} & \textbf{\textit{14.80\%}} \\
		\cline{2-6}
		& \multirow{3}{*}{20} & DAG~\cite{li2020structured} & 58.99 & 73.55 & 12.35\% \\
		\cline{3-6}
		& & Few-shot DAG & 54.94 & 55.00 & 9.37\% \\
		\cline{3-6}
		& & \textbf{\textit{Few-shot DAG + JS}} & \textbf{\textit{43.46}} & \textbf{\textit{47.22}} & \textbf{\textit{5.28\%}} \\
		\cline{2-6}
		& \multirow{3}{*}{50} & DAG~\cite{li2020structured} & 27.31 & 42.33 & 2.69\% \\
		\cline{3-6}
		& & Few-shot DAG & 23.32 & 28.64 & 1.07\% \\
		\cline{3-6}
		& & \textbf{\textit{Few-shot DAG + JS}} & \textbf{\textit{22.49}} & \textbf{\textit{27.29}} & \textbf{\textit{0.85\%}} \\
		
		\hline
	\end{tabular}
\end{table}

To manage the GPU memory consummation, the batch size is set as 1, 4, and 8 for experiments with 1, 5, and $>8$ training examples. Adam is used as the optimizer~\cite{duchi2011adaptive}, with learning rate set to $10^{-4}$, and is decayed by $0.96$ after every 10 epochs. The weight decay is $10^{-4}$. $w_1$ and $w_2$ are both set as 1. The backbone used is HRNET~\cite{sun2019deep}. $\alpha$ for updating the teacher model is set as:
\begin{equation}
    \alpha = \min\left(\max\left(1 - 1 / ({global}_{step} + 1), 0.99\right), 0.999\right),
\end{equation}
where ${global}_{step}$ is the global step of the training.

All input images are resized to $512 \times 512$ and normalized into an intensity range of $[0, 1]$. A rotation within the angle range of $30^\circ$, a scale within the range of $[0.8, 1.25]$, and a translation within the range of the landmark bounding box are applied as the data augmentation in the training. The noise added between the teacher and student model's input follows a $\mathcal{N}(0,0.1^2)$ distribution and the image intensity is clipped into $[0, 1]$ after adding the Gaussian noise.

\begin{figure}[t]
    \centering
    \includegraphics[width=1.0\textwidth]{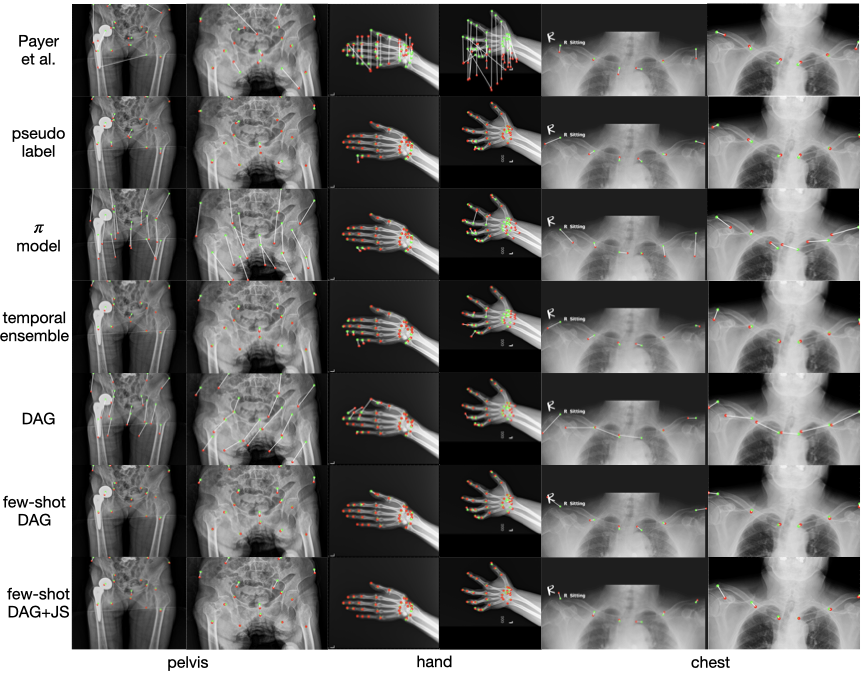}
    \caption{A visual illustration of two landmark localization results for the pelvis, hand and chest dataset. Green color indicates the \ac{GT}, red color indicates the prediction, while white line indicates the correspondence.}
    \label{fig:visual}
\end{figure}

\subsubsection{Comparison with other baselines}
\label{sec:comparison}

We compare against the fully-supervised \ac{DAG}~\cite{li2020structured} and also Payer \etal~\cite{payer2019integrating}, who introduced a heatmap based method focusing on leveraging the spatial information. Three semi-supervised methods are used for the comparison: (1) pseudo label~\cite{sohn2020simple}, which trains the \ac{SSL} model with the pseudo \ac{GT} generated by the \textit{pre-trained model} in Sec.~\ref{sec:dag}; (2) $\rm{\Pi}$-Model~\cite{laine2016temporal}, which maintains only a student model with the semi-supervised loss between the two input batches (one is with Gaussian noise and one is not); (3) temporal ensemble~\cite{laine2016temporal}, which updates the pseudo \ac{GT} after each epoch via the \ac{EMA} of historically and currently generated pseudo \ac{GT}. For the proposed method, performance in stages is offered, including few-shot \ac{DAG} and few-shot \ac{DAG} + \ac{JS}. It is worth mentioning that temporal ensemble consumes much longer training time than other methods.

Detailed results of validating the seven methods on the pelvis, hand and chest data set are presented in Tab. \ref{tab:comparison}. We can see that for the main evaluation metric, \ie{} mean and std Euclidean error, out of the fully-supervised methods, \ac{DAG} noticeably outperforms Payer \etal. For the semi-supervised methods, few-shot \ac{DAG} + JS outperforms $\rm{\Pi}$-Model with large margins while also outperforms pseudo label and temporal ensemble with notable margins. Furthermore, the proposed method shows consistent performance gains from \ac{DAG} to few-shot \ac{DAG}, and to few-shot \ac{DAG} + \ac{JS}, demonstrating the value of the proposed \ac{SSL} scheme and \ac{JS} divergence loss. For the supplementary evaluation metric - failure rate, similar trends can be observed. Even though few-shot \ac{DAG} + \ac{JS} does not achieve the lowest failure rate in one experiment, its failure rate is very close to the lowest value, \ie 1.31\% vs. 1.27\% on the pelvis.

\subsubsection{Ablation study on the scalability of few-shot \ac{DAG}}
\label{sec:ablation}

We show that the proposed few-shot \ac{DAG} can work well for different scales of training data. To illustrate this, we conduct experiments on varied numbers of labeled data. The corresponding results are shown in Tab.~\ref{tab:ablation}. We can see that the fully-supervised \ac{DAG} cannot converge on extremely few training examples, \ie 1 for pelvis, 1 for hand, 1 and 5 for chest, resulting in non-converged few-shot \ac{DAG} as well. On converged experiments with few training examples, for the main evaluation metric, the proposed \ac{SSL} scheme (\ac{DAG} vs. few-shot \ac{DAG}) achieves notable  performance improvements on all experiments. The proposed \ac{JS} divergence loss (few-shot \ac{DAG} vs. few-shot \ac{DAG} + \ac{JS}) performs best on most experiments, except one (hand-50) where the std error is comparable (17.55 vs. 17.51). For the supplementary evaluation metric, the proposed semi-supervised methods, including few-shot \ac{DAG} and few-shot \ac{DAG} + \ac{JS} fail much less than the fully-supervised method. While in most experiments, few-shot \ac{DAG} + \ac{JS} fails less than few-shot \ac{DAG}; only in three experiments (pelvis-10, pelvis -50, hand-50), comparable failure rates are achieved for semi-supervised \ac{DAG} with or without the \ac{JS} divergence loss.

\subsubsection{Visual results}
\label{sec:visual}

We select two examples for each data set and show the landmark localization results in Fig. \ref{fig:visual}. We can see that, Payer \etal{} and $\rm{\Pi}$-Model generally performs unstably while pseudo label and temporal ensemble can achieve reasonable landmark localization except for a few hard cases, \ie the second hand image. The fully-supervised \ac{DAG} can out-perform Payer \etal, however, it under-performs strong semi-supervised methods, \ie pseudo label and temporal ensemble. But, the proposed \ac{SSL} scheme (few-shot \ac{DAG}) and the \ac{JS} divergence loss (few-shot \ac{DAG} + \ac{JS}) can correct the less-optimal localization generated in fully-supervised \ac{DAG} and achieves the visually reasonable localization on all images including the hard cases.

\section{Conclusion}
\label{sec:conclusion}

In this paper, we introduced few-shot \ac{DAG}, an \ac{SSL} enhancement of \ac{DAG} that significantly improves landmark localization. It first trains a \ac{DAG} model on a few labeled training examples (\eg five), and then fine-tunes the trained model on a large number of unlabeled training examples using consistency losses tailored for \ac{CNN} and \ac{GCN} outputs. Overall, our approach achieves strong landmark localization performances with only a few training examples. As shown in the validation on three datasets, the proposed few-shot DAG consistently out-performs both previous fully-supervised and semi-supervised methods with notable margins, indicating its good performance, robustness and potentially wide application in the future.

\bibliographystyle{splncs04}
\bibliography{refs}

\end{document}